\documentclass{bmvc2k}

\usepackage{multirow}
\usepackage{amssymb}

\title{Motion-Aware Feature for Improved Video Anomaly Detection}

\addauthor{Yi Zhu}{yzhu25@ucmerced.edu}{1}
\addauthor{Shawn Newsam}{snewsam@ucmerced.edu}{1}
\addinstitution{
 Electrical Engineering and Computer Science\\
 University of California, Merced\\
 Merced, US
}
\runninghead{Yi Zhu, Shawn Newsam}{}

\begin{document}

\maketitle

\begin{abstract}
Motivated by our observation that motion information is the key to good anomaly detection performance in video, we propose a temporal augmented network to learn a motion-aware feature. This feature alone can achieve competitive performance with previous state-of-the-art methods, and when combined with them, can achieve significant performance improvements. Furthermore, we incorporate temporal context into the Multiple Instance Learning (MIL) ranking model by using an attention block. The learned attention weights can help to differentiate between anomalous and normal video segments better. With the proposed motion-aware feature and the temporal MIL ranking model, we outperform previous approaches by a large margin on both anomaly detection and anomalous action recognition tasks in the UCF Crime dataset. 
\end{abstract}

%-------------------------------------------------------------------------
\section{Introduction}
\label{sec:intro}

Anomaly detection in video is one of the long standing problems in computer vision and has extensive applications in surveillance monitoring, such as detecting illegal activities, traffic accidents and unusual events etc. Millions of surveillance cameras are being deployed in public places worldwide. However, most of the cameras are just passively recording without actually having any monitoring capability. With petabytes of data generated by the video cameras every minute, it is not possible to understand this large corpus of video data through human effort. We need machine vision to automatically detect anomalies within a video.

Recognizing anomaly in unconstrained videos is extremely hard. The challenges include insufficient annotated data due to the rare occurrence of anomalies, large inter/intra class variations, subjective definition of anomalous events, low resolution of surveillance videos, etc. As humans, we recognize anomalies using our common sense. 
% if a person is running adversely to the traffic, there maybe an anomaly. 
For example, if multiple people crowd in a street that usually has less traffic, there maybe an anomaly. If violent events such as fighting happen, there maybe an anomaly. For machines, they don't have common sense but only have visual features. In general, the stronger the visual features, the better the anomaly detection performance is expected. In this work, we demonstrate how to obtain strong visual features by incorporating motion information.

Previous work \cite{Cheng_2015_CVPR,Yang_2015_ICCV,Shao_2016_CVPR,Narasimhan_2018_MTA,Zhang_Relationship_2019_AAAI} uses either hand crafted features or deep learned features to detect anomalies. Since their performance are reported on different datasets, we conduct an experiment here to make a fair comparison among these features. We evaluate on the UCF Crime dataset \cite{Sultani_Crime_2018_CVPR}, a recently released large-scale real world video anomaly benchmark. We adopt the Multiple Instance Learning (MIL) framework proposed in \cite{Sultani_Crime_2018_CVPR} to report the corresponding Area Under the receiver operating characteristic Curve (AUC), while changing only the input features. 
For volume-based features such as C3D \cite{Tran_C3D_2015_ICCV} and I3D \cite{Carreira_I3D_2017_CVPR}, the input to the network is a 16-frame video clip. For image-based features such as VGG16 \cite{Simonyan_VGG_2015_ICLR} and Inception \cite{Szegedy_Inception_2015_CVPR}, we input the same 16-frame video clip as a mini-batch to the network and average the features. As can be seen in Table \ref{tab:feature_comparison}, we have an important observation that volume-based features that incorporate motion information perform much better than image-based features, regardless of network depth and feature dimension. This intuitively makes sense because most anomalies are irregular abrupt motion patterns, and motion-aware features should be more suitable to detect such events. 

\begin{table}
\begin{center}
\begin{tabular}{|c|c|c|c|c|}
\hline
Features & Network & Dimension & Motion & AUC ($\%$) \\
\hline\hline
% SIFT & hand crafted &  & $61.2$ \\
% IDT  & hand crafted & \checkmark & $67.9$ \\
% \hline
VGG16 \cite{Simonyan_VGG_2015_ICLR} & deep & 4096 & & $68.7$ \\
C3D \cite{Tran_C3D_2015_ICCV} & deep & 4096 &\checkmark & $74.4$ \\
\hline
Inception \cite{Szegedy_Inception_2015_CVPR}  & very deep & 1024 & & $69.2$ \\
I3D \cite{Carreira_I3D_2017_CVPR} & very deep & 1024 & \checkmark & $75.4$ \\
\hline
\end{tabular}
\end{center}
\caption{Evaluation of different features on the UCF Crime dataset \cite{Sultani_Crime_2018_CVPR}. \textit{Motion} indicates whether temporal information is involved. We observe that features incorporating motion information (C3D and I3D) perform much better than features extracted from individual images (VGG16 and Inception), regardless of network depth and feature dimension. }
\label{tab:feature_comparison}
\vspace{-2ex}
\end{table}

Motivated by the above observation, our goal is to learn a strong visual feature by incorporating as much temporal information as we can from the raw video frames. In this work, we propose a temporal augmented network to learn motion-aware features in an unsupervised manner. Our learned feature is efficient to compute, and shown to be competitive with other deep learned features such as C3D \cite{Tran_C3D_2015_ICCV}. When combined with other features, we obtain a significant performance improvement. Our contributions are as below.

\begin{itemize}
    \item We propose a temporal augmented network to learn motion-aware features. Such features are shown to be complementary to existing features. 
    \item We introduce an attention-based temporal MIL ranking model, which can take temporal context into the picture and differentiate between anomalous and normal events better.
    \item We compare with and outperform several state-of-the-art approaches on both anomaly detection and anomalous action recognition tasks in the UCF Crime dataset.
\end{itemize}

\section{Related Work}
\label{sec:related}

Here, we discuss additional work related to ours, focusing mainly on the temporal modeling of videos. Video is more than just a stack of images. Modeling the temporal relationship among frames can help understanding the video better. Initial attempts use tracking to design hand crafted features, such as IDT \cite{Wang_IDT_2013_ICCV}. Recent deep learning based methods use temporal convolution \cite{Varol_2017_TemporalConv_PAMI}, 3D convolution \cite{Tran_C3D_2015_ICCV,Carreira_I3D_2017_CVPR}, temporal segment networks \cite{Wang_2016_TSN_ECCV}, two-stream networks \cite{Simonyan_2014_TwoStream_NeurIPS} etc. Among them, two-stream based approaches using optical flow are the top performers on most video benchmarks.

In this work, instead of directly using optical flow, we propose a temporal augmented network as an autoencoder to learn a compact motion-aware feature. This feature is generic, efficient and can be easily integrated with other methods using early fusion. We also incorporate temporal context into classical MIL ranking models by using an attention mechanism.  
The most similar literature to ours is \cite{Xu_Anomaly_2015_BMVC,Sultani_Crime_2018_CVPR}, however, there are several differences. \cite{Xu_Anomaly_2015_BMVC} experiments with small scale datasets which does not require MIL formulation, while ours introduces a temporal MIL framework with an attention module. \cite{Sultani_Crime_2018_CVPR} serves as a baseline where we show substantial improvements by our proposed techniques. We fully exploit the temporal constraints within a video for improved anomaly recognition and detection. At the same time, our whole framework runs faster than real-time, which  makes it directly applicable to real world problems. 

\section{Methodology}
\label{sec:method}

\subsection{Problem Formulation}
Given a long untrimmed video, we want to know whether it contains an anomalous event and where the event happens. Due to the massive amount of video recordings and the rare occurrence of anomalies, it is very challenging and costly to obtain precise frame-level annotations to train a powerful neural network. Most video anomaly detection datasets \cite{Sultani_Crime_2018_CVPR,Rabiee_2016_crowd_AVSS} only provide video-level labels. Hence, in this work, we need to develop a weakly supervised approach using such datasets. Our goal is to learn a regressor that can predict the anomaly score for a video clip and detect possible anomalous event within a video. 

\begin{figure*}[t]
	\centering
	\includegraphics[trim={0 0 0 0},clip,height=3.2cm,width=0.8\linewidth]{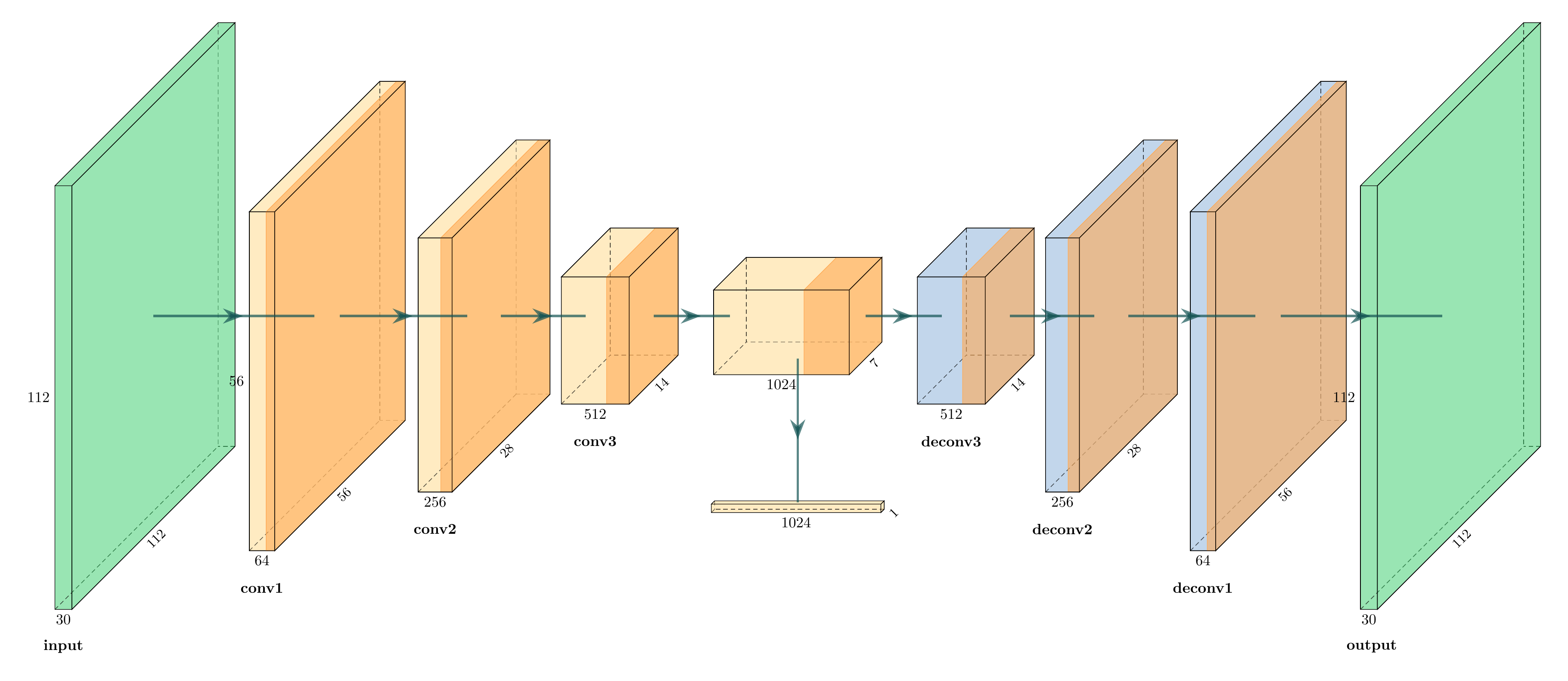}
	\vspace{-2ex}
	\caption{Temporal augmented network. The input (green) is a stack of 15 optical flow maps which this network aims to reconstruct by learning a compact representation. We then use a global average pooling operation to derive our $1024 \times 1$ motion-aware feature. }
	\label{fig:tan}
	\vspace{-2ex}
\end{figure*}

\subsection{Temporal Augmented Network}

As we know, volume-based features such as C3D and I3D are computed on multiple video frames using 3D convolutions. They already contain temporal information. That is the reason they outperform image-based features such as VGG16 and Inception. However, as shown in recent action recognition literature \cite{Wang_2016_TSN_ECCV,Carreira_I3D_2017_CVPR,Xie_2018_S3D_ECCV}, volume-based features alone cannot achieve state-of-the-art performance. Combining them with optical flow, as in the popular two-stream network \cite{Simonyan_2014_TwoStream_NeurIPS}, performs the best on most video classification benchmarks. This indicates that learning spatiotemporal features directly from raw video frames is challenging. Extra motion information such as optical flow can help. 

Motivated by this observation, we want to learn a motion-aware feature that can complement existing features for improved video anomaly detection. 
In this work, we propose a temporal augmented network as shown in Figure \ref{fig:tan}. The network is an autoencoder. Its input is some prior motion information pre-computed from raw video frames, such as optical flow. This forces the network to directly learn complex motion patterns. Then we aim to encode a compact representation so that we can use it to recover the input as closely as possible. This representation is our motion-aware feature and can be used to detect video anomalies. 

Since optical flow is the most widely adopted motion representation, we use it as the input to the autoencoder. Specifically, we use the state-of-the-art neural network-based flow estimator PWCNet \cite{Sun_2018_PWCNet_CVPR} to compute the optical flow between adjacent frames. We also compare several other motion representations in Section \ref{sec:discussion}. Similar to C3D, we choose $16$ frames as a video clip $\mathcal{V}$ and resize them to a resolution of $112 \times 112$. We then compute the optical flow on the resized frames. Each optical flow map has two channels, one for horizontal movement and the other for vertical. Hence, the final input to our temporal augmented network is a stack of $15$ optical flow maps $\mathcal{F}$ with the dimension of $30 \times 112 \times 112$.  

Bearing efficiency in mind, we design the temporal augmented network to have only 7 layers: 3 encoder layers, 1 bottleneck layer and 3 decoder layers. All layers consist of a 2D convolutional layer followed by a ReLU activation. We use a stride of 2 to halve the feature map resolution instead of pooling. The network can be trained in an unsupervised manner on the target dataset using L1 per-pixel reconstruction loss, 
\begin{equation}
    \text{loss}_{\; \text{recon}} = | \mathcal{F} - \tilde{\mathcal{F}} | 
\end{equation}
where $\tilde{\mathcal{F}}$ is the reconstructed flow map. Once the training is completed, we can treat it as a feature extractor. For each video clip with $16$ frames, we perform a forward pass until the bottleneck layer and conduct a global average pooling operation to derive a $1024\times1$ feature. This will be our motion-aware feature for anomaly detection. If we want to use it with other features, we can simply concatenate them together. Note that the motion-aware feature is learned from optical flow, hence it contains only motion information without looking at the original frame pixels. We don't perform spatiotemporal feature learning. This will help the network to focus on the moving parts and learn appearance-invariant features.

\begin{figure*}[t]
	\centering
	\includegraphics[trim={0 0 0 0},clip,width=1.0\linewidth]{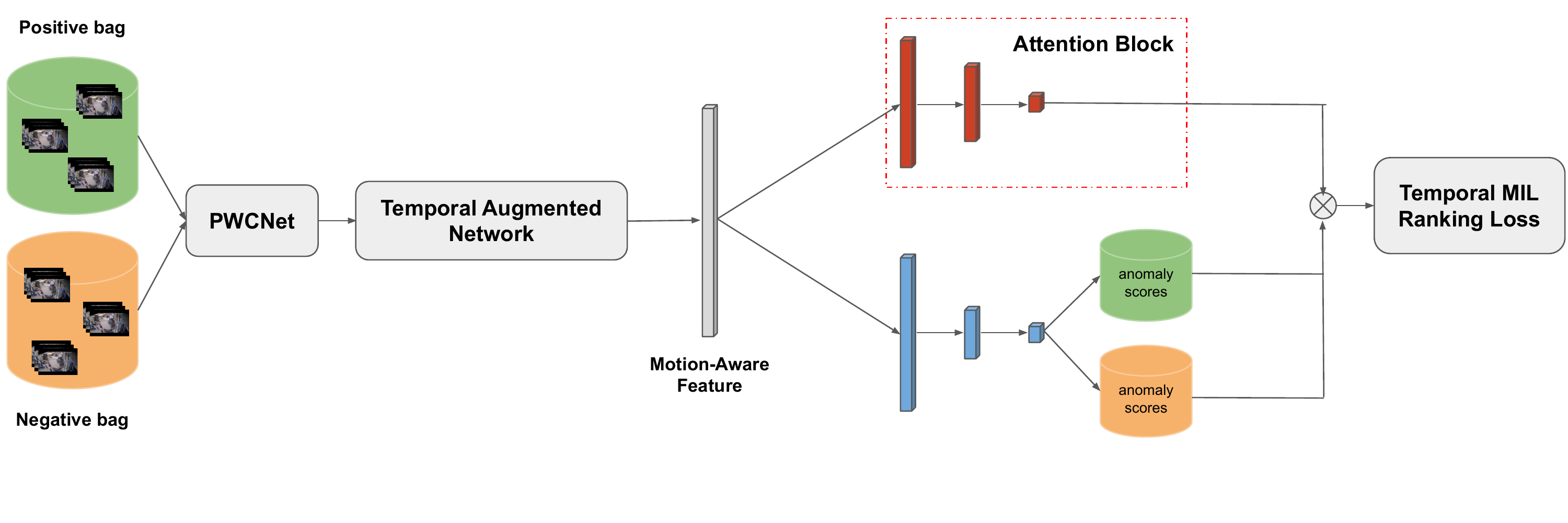}
	\vspace{-6ex}
	\caption{Overall framework. We first obtain the motion-aware feature and then compute the predicted anomaly scores. The attention block is used together with the proposed temporal MIL ranking loss to incorporate temporal context into training for better anomaly detection.}
	\label{fig:framework}
	\vspace{-2ex}
\end{figure*}

\subsection{Attention-based Temporal MIL Ranking Model}

\paragraph{MIL Formulation}
Since the precise temporal locations of anomalous events in videos are unknown, we cannot simply learn anomaly patterns like in a standard classification problem. Instead, we can treat it as a Multiple Instance Learning (MIL) problem. 

In our scenario, we only have video-level annotations. A video containing anomalies is labeled as positive and a normal video is labeled as negative. Following \cite{Sultani_Crime_2018_CVPR}, we represent a positive video as a positive bag $\mathcal{B}_{a}$, where different temporal segments are individual instances in the bag, $(a^{1}, a^{2}, . . . , a^{m})$, where $m$ is the number of instances in the bag. We assume that at least one of these instances contains the anomaly. Similarly, the negative video is denoted by a negative bag, $\mathcal{B}_{n}$, where temporal segments in this bag are negative
instances $(n^{1}, n^{2}, . . . , n^{m})$. In the negative bag, none of the instances contain an anomaly. In this work, we divide each video into a fixed number of segments (e.g., $32$ segments) during training. These segments of a video are the instances in a bag.

\paragraph{MIL Ranking Model}

Following previous work \cite{Sultani_Crime_2018_CVPR}, we formulate anomaly detection as an anomaly score regression problem. We hope the segments from an anomalous video to have higher anomaly scores than the segments from a normal video. If we have the segment-level annotations, we can simply use a ranking loss as 
\begin{equation}
    f(\mathcal{V}_{a}) > f(\mathcal{V}_{n}), 
\end{equation}
where $\mathcal{V}_{a}$ and $\mathcal{V}_{n}$ are anomalous and normal video segments. \textit{f} is the function that maps a video segment to its corresponding predicted anomaly scores ranging from 0 to 1. Here, \textit{f} is designed to be a 3-layer fully-connected neural network. The first fully-connected layer has $512$ units followed by $32$ unit and $1$ unit fully-connected layers. Dropout regularization is used between these layers. We use ReLU activation and Sigmoid activation for the first and the last fully-connected layers, respectively.
However, we only have access to video-level annotations. \cite{Sultani_Crime_2018_CVPR} thus proposed a MIL ranking loss
\begin{equation}
    \max_{i \in \mathcal{B}_{a}}f(\mathcal{V}_{a}^{i}) > \max_{i \in \mathcal{B}_{n}}f(\mathcal{V}_{n}^{i}). 
    \label{eq:ranking_loss}
\end{equation}
Here, $max$ is taken over all video segments in each bag. The intuition behind this ranking objective is that the segment with highest anomaly score in the positive bag should rank higher than the segment with highest anomaly score in the negative bag because a negative bag does not contain any anomaly. In order to keep a large margin between the positive and negative instances, \cite{Sultani_Crime_2018_CVPR} introduced a hinge-based ranking loss 
\begin{equation}
    l(\mathcal{B}_{a}, \mathcal{B}_{n}) = \text{max} (0, 1 - \max_{i \in \mathcal{B}_{a}}f(\mathcal{V}_{a}^{i}) + \max_{i \in \mathcal{B}_{n}}f(\mathcal{V}_{n}^{i})). 
\label{eq:hinge_ranking_loss}
\end{equation}

However, there are at least two limitations to this ranking loss. First, we note that Equation \ref{eq:ranking_loss} ignores the underlying temporal structure of the anomalous video. A single max operation is not expressive. There maybe an anomalous video that contains multiple anomaly events. For a normal video, some segments could also look anomalous. Reasoning on temporal context should be useful to differentiate anomalous and normal video segments better. Second, the hinge-based ranking loss in Equation \ref{eq:hinge_ranking_loss} can easily lead to a degenerate solution where we predict most video segments to be normal.  

\paragraph{Temporal MIL Ranking Model}

In this section, we introduce our temporal MIL ranking model by using temporal context information. Motivated by the above limitations, we turn to an attention-based framework which can capture the total anomaly score of a video, 
\begin{equation}
    \sum_{i \in \mathcal{B}_{a}} w_{i} f(\mathcal{V}_{a}^{i}) > \sum_{i \in \mathcal{B}_{n}} w_{i} f(\mathcal{V}_{n}^{i}). 
    \label{eq:weighted_ranking_loss}
\end{equation}
where $w_{i}$ indicates the learned attention weights. The intuition is, the overall anomaly score of an anomaly video should be larger than that of a normal video. We should include temporal context into consideration and compute the anomaly score video-wise, not segment-wise. 

The attention weights are learned end-to-end within the network. As can be seen in Figure \ref{fig:framework}, we add an attention block after the input features. The block consists of three fully-connected layers and two tanh activations in between. The first fully-connected layer has $256$ units followed by $64$ unit and $1$ unit fully-connected layers. For each video with \textit{m} segments, we will learn a $1 \times m$ attention score for all the segments. Similar to Equation \ref{eq:hinge_ranking_loss}, our hinge-based temporal ranking loss is defined as 
\begin{equation}
    l(\mathcal{B}_{a}, \mathcal{B}_{n}) = \text{max} (0, 1 - \sum_{i \in \mathcal{B}_{a}} w_{i} f(\mathcal{V}_{a}^{i}) + \sum_{i \in \mathcal{B}_{n}} w_{i} f(\mathcal{V}_{n}^{i})).
\label{eq:hinge_temporal_ranking_loss}
\end{equation}

We also employ the sparsity constraints \cite{Sultani_Crime_2018_CVPR,Zhang_Graph_2019_CVPR} because anomalies occur only rarely. There should be only a few segments that have high anomaly score. In the end, our final loss function becomes 
\begin{equation}
    \text{Loss} = l(\mathcal{B}_{a}, \mathcal{B}_{n}) + \lambda_{1} \sum_{i \in \mathcal{B}_{a}} w_{i} f(\mathcal{V}_{a}^{i}).
\label{eq:final_loss}
\end{equation}
$\lambda_{1}$ is the loss weight for the sparsity constraint. Note that we do not use the temporal smoothness constraint introduced in \cite{Sultani_Crime_2018_CVPR}. We empirically find it harmful for model training. Our overall framework can be seen in Figure \ref{fig:framework}.

\begin{table}[t]
    \begin{minipage}{0.5\textwidth}%
	\centering
	\subfigure{
		\scalebox{0.9}{
			\begin{tabular}{| c | c |}
				\hline
				Method	&  AUC ($\%$)		\\
				\hline	
				Hasan et al. \cite{Hasan_2016_CVPR} & $50.6$ \\
				Lu et al. \cite{Lu_2013_150fps_ICCV} & $65.5$ \\
				Sultani et al. \cite{Sultani_Crime_2018_CVPR} & $75.4$ \\
				\hline	
				MA & $72.1$ \\
				Hasan et al. \cite{Hasan_2016_CVPR} + MA & $62.7$ \\
				Lu et al. \cite{Lu_2013_150fps_ICCV} + MA & $73.4$ \\
				Sultani et al. \cite{Sultani_Crime_2018_CVPR} + MA & $\mathbf{79.0}$ \\
				\hline
			\end{tabular}
		}
	}
    \end{minipage}%
    %\qquad
    \begin{minipage}{0.5\textwidth}%
	    \centering
	   \ subfigure{
		    \scalebox{1.0}{
            \includegraphics[width=5.6cm]{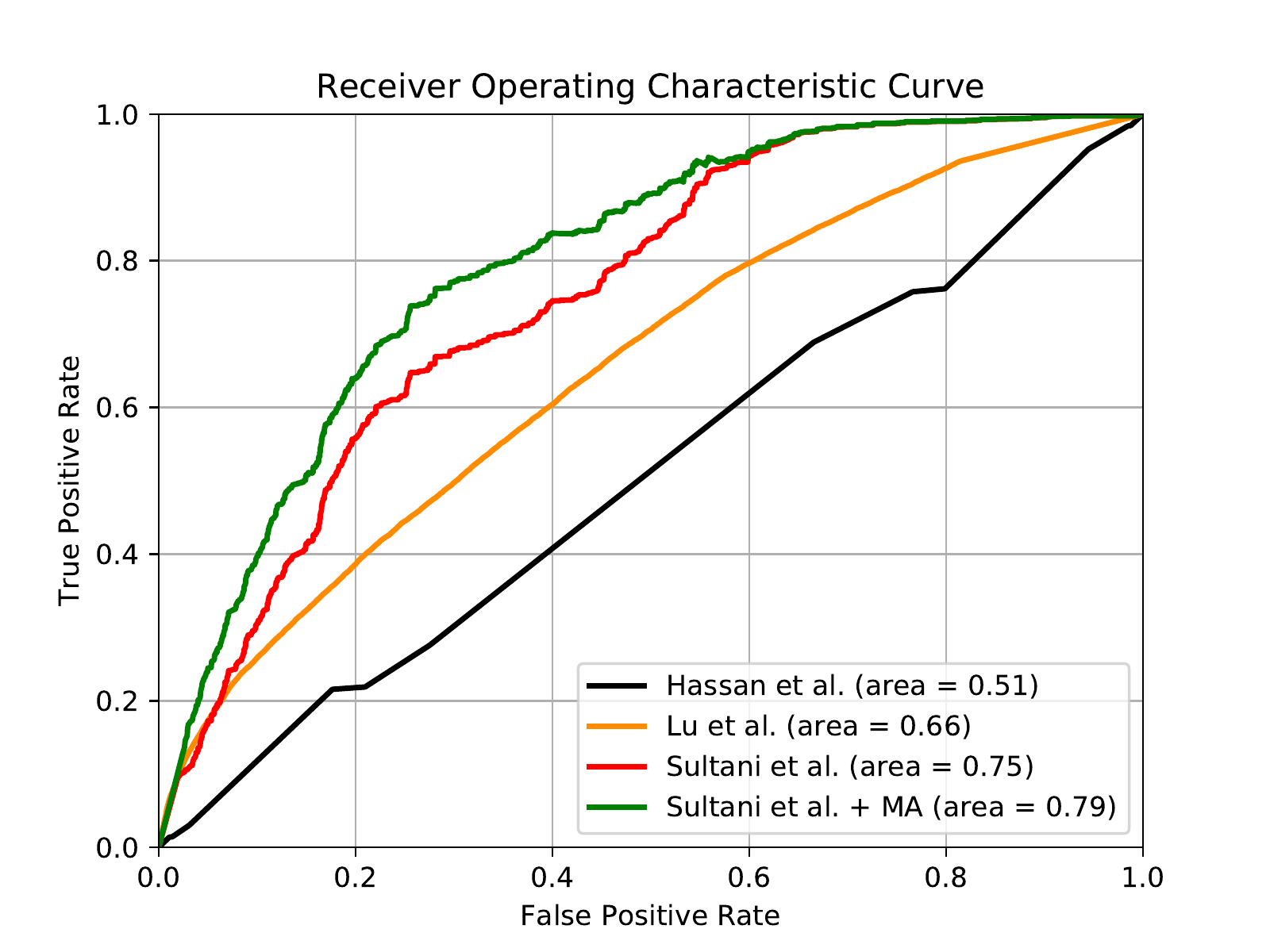}
            }
        }
    \end{minipage}%
%     \begin{minipage}{0.5\textwidth}%
% 	\centering
% 	\subfloat{
% 	    \scalebox{1.0}{
% 	        \begin{tabular}{| c | c |}
% 				\hline
% 				Method	&  AUC ($\%$)		\\
% 				\hline	
% 				SIFT + MA & $73.6$ \\
%                 IDT + MA & $74.1$ \\
%                 VGG16 + MA & $74.5$ \\
%                 C3D + MA & $78.7$ \\
%                 Inception + MA & $75.1$ \\
%                 I3D + MA & $79.6$ \\
%                 \hline
% 		    \end{tabular}
%         }
%     }
%     \end{minipage}%
    % \vspace{2ex}
	\caption{Performance comparison on the UCF Crime dataset. \textit{MA} indicates our motion-aware feature learned from temporal augmented network.  Left: Comparison to state-of-the-art approaches. Our motion-aware feature complements existing methods for better anomaly detection. Right: Visual comparison in terms of ROC and AUC. \cite{Sultani_Crime_2018_CVPR} with motion-aware feature (green) achieves higher true positive rates than without (red). \label{tab:sota}}%
	\vspace{-2ex}
\end{table}

\section{Experiments}
\label{sec:experiments}

\subsection{Dataset}

Previous datasets \cite{Lu_2013_150fps_ICCV,Li_2014_UCSDPed_PAMI,Rabiee_2016_crowd_AVSS} for video anomaly detection are either small in terms of the number of videos or have limited anomaly classes. 
%For example, the UCSD Ped2 dataset \cite{Li_2014_UCSDPed_PAMI} only has $28$ videos and the videos are all captured at a single location. This does not reflect real-life situations and may lead to biased results. 
Since we are doing comparisons among multiple features, we need a large, diverse and balanced dataset to reach a convincing conclusion. We use a recently released large-scale real world anomaly detection benchmark, UCF Crime \cite{Sultani_Crime_2018_CVPR}, to evaluate our model and design choices. 
This dataset consists of $1900$ real-world surveillance videos, half of which contain anomalous events and the other half normal activities. For the anomalous videos, there are $13$ different classes, including \textit{Abuse, Arrest, Arson, Assault, Accident, Burglary, Explosion, Fighting, Robbery, Shooting, Stealing, Shoplifting, and Vandalism}. The official training split divides the dataset into two parts: the training set consisting of $800$ normal and $810$
anomalous videos and the testing set including the remaining $150$ normal and $140$ anomalous videos. Following previous works \cite{Sultani_Crime_2018_CVPR}, we use frame based receiver operating characteristic (ROC) curves and corresponding area under the curve (AUC) to evaluate the performance of our method.

\subsection{Implementation Details}

We use the PyTorch framework to train our model. 
For the temporal augmented network, we randomly select video clips of $16$ frames and use PWCNet \cite{Sun_2018_PWCNet_CVPR} to compute the optical flow. 
% PWCNet  is an efficient network which can process video frames at $112 \times 112$ resolution at a speed of $700$+ frame per second (fps). 
The batch size is set to $50$. We use the Adagrad optimizer with an initial learning rate of $0.005$. We train the model for a total of $50$K iterations, and decrease the learning rate by half at $25$K, $40$K and stop at $50$K. 
For the MIL ranking model, we first divide each video into $32$ non-overlapping segments. If the video has less than $32$ frames, we duplicate its frames. Within each segment, we compute our motion-aware feature for every non-overlapping 16-frame video clip. If the segment has multiple 16-frame video clips, we take the average of all features followed by a L2 normalization. Hence, for each video, we have a $32 \times 1024$ feature. To train the MIL ranking model, we randomly select $30$ positive and $30$ negative bags as a mini-batch. We use the Adagrad optimizer with an initial learning rate of $0.001$. We train the model for a total of $10$K iterations, and decrease the learning rate by half at $4$K, $8$K and stop at $10$K. $\lambda_{1}$ is set to $8\times10^{-5}$. For all other features used in this paper such as C3D and I3D, we adopt the implementation kindly provided by the original authors \cite{Tran_C3D_2015_ICCV,Carreira_I3D_2017_CVPR}.

\begin{figure*}[t]
	\centering
	\includegraphics[trim={0 0 0 0},clip,width=1.0\linewidth]{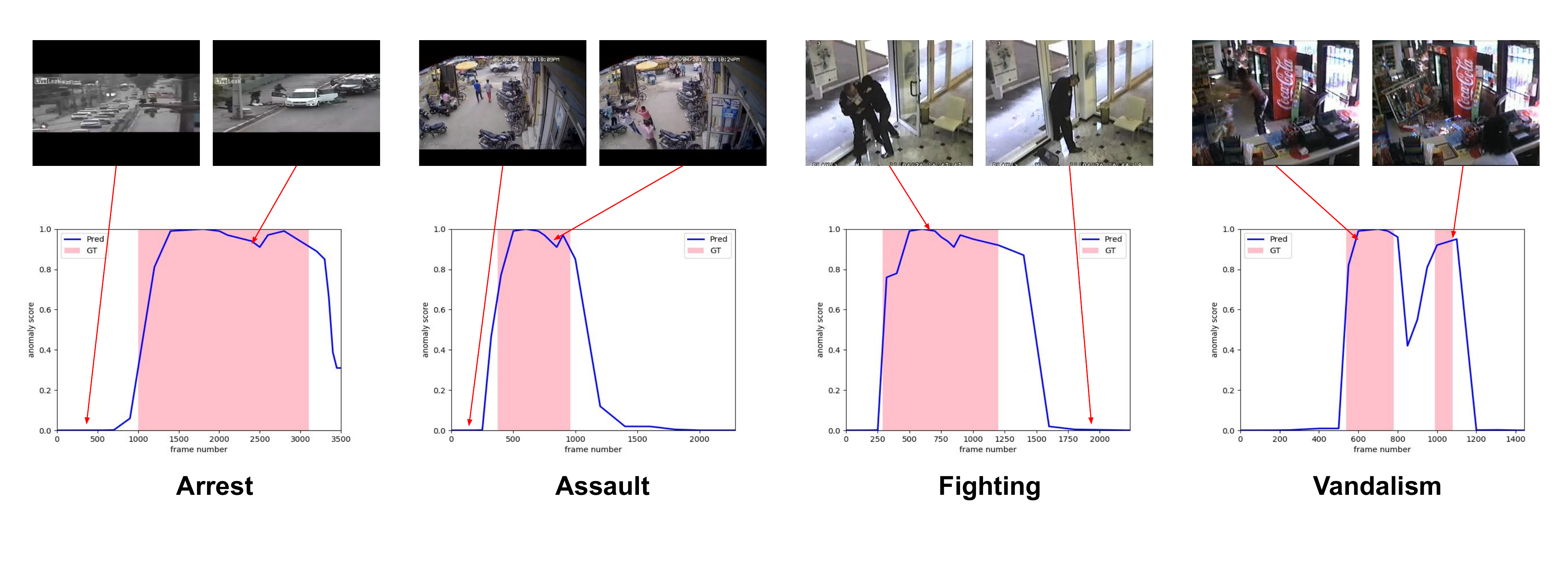}
	\vspace{-8ex}
	\caption{Visual examples of prediction results. For the anomalous frames, our model is able to provide accurate detection by generating high anomaly scores. For the normal frames, our model consistently produces low anomaly scores. }
	\label{fig:samples}
	\vspace{-2ex}
\end{figure*}

\subsection{Results}
We present our results in Table \ref{tab:sota}. We compare our method with a state-of-the-art approach \cite{Sultani_Crime_2018_CVPR} and two other baselines \cite{Lu_2013_150fps_ICCV,Hasan_2016_CVPR} for anomaly detection. In order to make fair comparisons, we keep the model training setting the same. 

As we can see in Table \ref{tab:sota} left, our motion-aware feature learned from the temporal augmented network achieves competitive performance with the previous best \cite{Sultani_Crime_2018_CVPR} in terms of anomaly detection AUC score ($72.1$ VS $75.4$), but has smaller size (1024-dim VS 4096-dim) and faster speed (400+ fps VS 300+ fps). When combined with \cite{Sultani_Crime_2018_CVPR}, we can achieve a performance improvement of $3.6\%$ ($75.4 \rightarrow 79.0$). As for per-class breakdown, we observe that classes with fast motion benefit a lot from our motion-aware feature. For example, Arrest ($46.0 \rightarrow 57.4$), Assault ($20.8 \rightarrow 41.2$) and Fighting ($32.4 \rightarrow 47.1$).
Similarly, when combined with \cite{Lu_2013_150fps_ICCV,Hasan_2016_CVPR}, we are able to get significant performance boosts of $12.1\%$ and $7.9\%$, respectively. 
This demonstrates the effectiveness of our learned motion-aware feature. 

In terms of visualization, we show the comparison of ROC curves in Table \ref{tab:sota} right. We can see that \cite{Sultani_Crime_2018_CVPR} with our motion-aware feature (green) achieves higher true positive rates than without (red) at low false positive rates. This will help to reduce the false alarm rate. 

We also combine our motion-aware feature with other widely adopted features such as VGG16, Inception and I3D. We observe consistent improvements: VGG16 ($68.7 \rightarrow 74.2$), Inception ($69.2 \rightarrow 74.9$) and I3D ($75.4 \rightarrow 79.8$). The large improvements indicate the strong complementarity of our feature. At the same time, we may conclude that motion patterns are strong indicators for detecting anomalies. The more motion information captured in the visual feature, the better performance we will have. 

In Figure \ref{fig:samples}, we show several visual examples of our qualitative results. We can see that for the anomalous frames, our model is able to provide successful and
timely detection by generating high anomaly scores. For the normal frames in which no anomaly occurs, our model consistently produces low (almost zero) anomaly scores.  

\begin{table}[t]
    \hspace{1ex}
    \begin{minipage}{0.35\textwidth}%
	\centering
	\subfigure{
		\scalebox{0.8}{
			\begin{tabular}{| c | c |}
				\hline
				Method	&  AUC ($\%$)		\\
				\hline	
				MA (max) & $70.6$ \\
				MA (attention) & $72.1$ \\
				\hline
				\cite{Sultani_Crime_2018_CVPR} (max) & $75.4$ \\
				\cite{Sultani_Crime_2018_CVPR} (attention) & $76.2$ \\
				\hline
				\cite{Sultani_Crime_2018_CVPR} + MA (max) & $77.1$ \\
				\cite{Sultani_Crime_2018_CVPR} + MA (attention) & $79.0$ \\
				\hline
			\end{tabular}
		}
	}
    \end{minipage}%
    % \qquad
    \begin{minipage}{0.7\textwidth}%
	    \centering
	    \subfigure{
		    \scalebox{1.0}{
            \includegraphics[width=6.6cm]{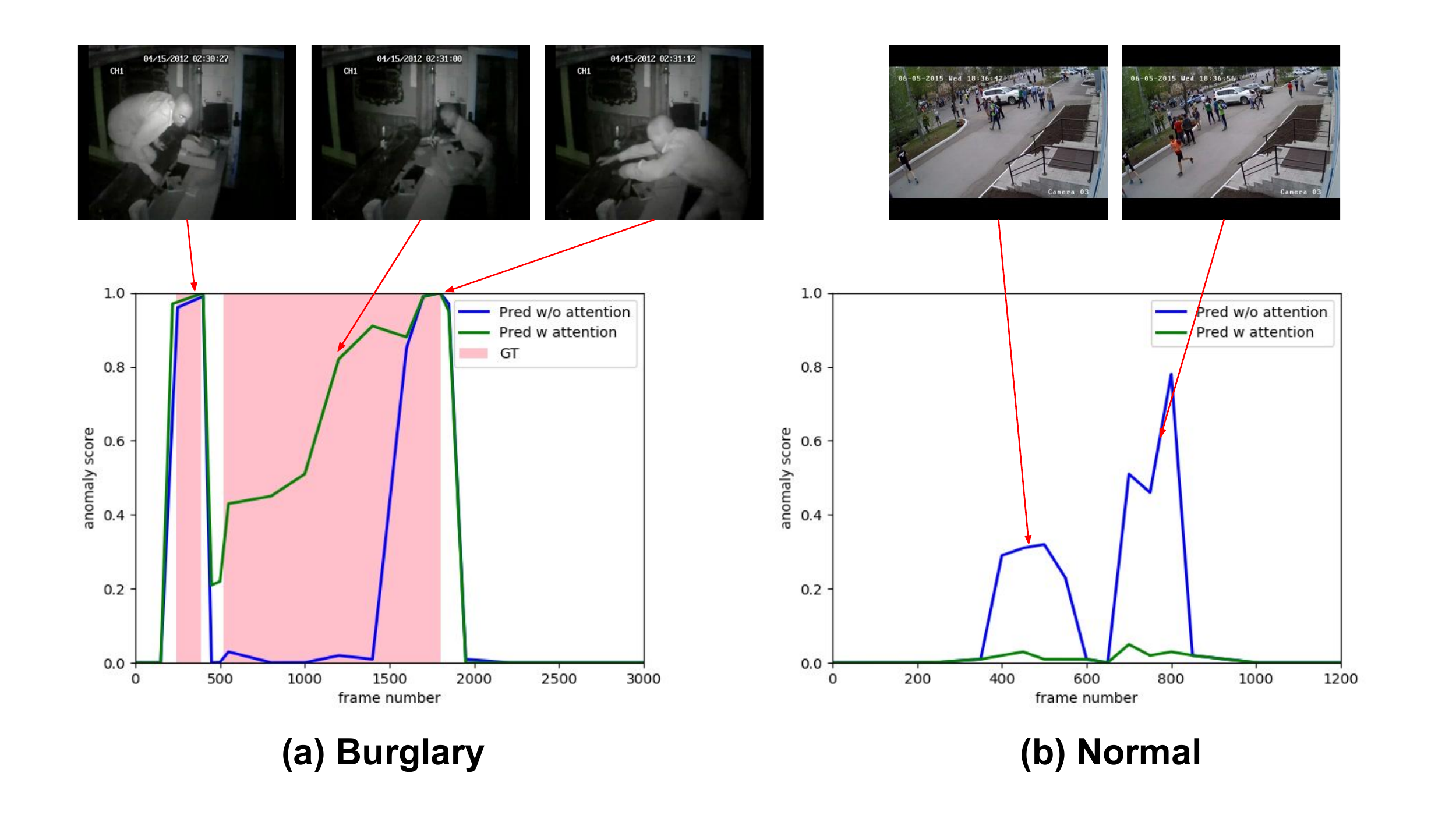}
            }
        }
    \end{minipage}%
    % \vspace{-2ex}
	\caption{Attention is useful. Left: Quantitative results. Right: Two visual examples. Temporal context can help to differentiate between anomalous and normal events better. \label{tab:discussion}}%
	\vspace{-4ex}
\end{table} 

\vspace{-2ex}
\section{Discussion}
\label{sec:discussion}

\subsection{Effectiveness of Attention Mechanism}

In this section, we investigate the effectiveness of the attention mechanism in temporal MIL ranking model, to see the benefit of using Equation \ref{eq:hinge_temporal_ranking_loss} over Equation \ref{eq:hinge_ranking_loss}. 
As can be seen in Table \ref{tab:discussion} left, adding attention consistently brings us $1\%$ to $2\%$ AUC improvement. We believe temporal context plays a key role here to differentiate anomalous and normal events. In terms of visualization, we show two examples in Table \ref{tab:discussion} right. The first video contains burglary events. Without attention, the model (blue) fails to report the anomaly from frame 500 to 1600 which looks normal. After adding attention, our method can detect the anomalous event there (green) because it has knowledge of the temporal context. 
%Under the context, frame 500 to 1600 gets much higher predicted anomaly scores. 
The second video doesn't contain any anomalies but there are people grouping and running in the middle. Without attention, the model classifies the middle two parts as anomalies (high spike of blue curve). After incorporating attention, the model stops producing high anomaly scores for those parts (green).

\subsection{Ablation Study on Motion Representations}

Recall from Section \ref{sec:method} that any motion representation can be fed to our temporal augmented network as input. There are many flow estimators, such as TVL1, FlowFields, FlowNet2 and PWCNet  etc. Besides optical flow, we also have other motion representations, such as motion vectors and video saliency etc. Here, we perform an ablation study among these representations to see which one is the most effective. 

First, we compare different flow estimators. TVL1 \cite{Zach_2014_TVL1_DAGM} and FlowFields \cite{Bailer_2015_FlowFields_ICCV} are classical methods, while FlowNet2 \cite{Ilg_2017_FlowNet2_CVPR} and PWCNet \cite{Sun_2018_PWCNet_CVPR} are neural network based methods. As can be seen in Table \ref{tab:motion} left, FlowNet2 achieves the best AUC score due to its accurate and sharp flow predictions. However, it is relatively slow to compute. 
% and thus not ready to be used in the application of video anomaly detection that requires real-time processing. 
PWCNet is a good trade-off, performing competitively to FlowNet2 but running significantly faster. 

Second, we compare different types of motion representations: motion vectors, optical flow and video saliency. Here, we use PWCNet \cite{Sun_2018_PWCNet_CVPR} to compute optical flow, and we use a state-of-the-art method \cite{Jiang_2018_Saliency_ECCV} to obtain video saliency. 
As can be seen in Table \ref{tab:motion} left, PWCNet achieves the best performance. Both the motion vectors and video saliency perform badly. We find that the resolution of motion vectors is too coarse to extract useful motion information. For video saliency, we observe that the predictions are not consistent across frames thus may complicate the learning process of temporal augmented network. 

\begin{table}[t]
    \begin{minipage}{0.5\textwidth}%
	\centering
	\subfigure{
		\scalebox{0.9}{
			\begin{tabular}{| c | c | c |}
				\hline
				Method	&  Speed (fps)  &  AUC ($\%$)		\\
				\hline	
				TVL1 \cite{Zach_2014_TVL1_DAGM}$^{*}$ &  $102$ &$71.7$ \\
				FlowFields \cite{Bailer_2015_FlowFields_ICCV} & $3$  & $70.6$ \\
				FlowNet2 \cite{Ilg_2017_FlowNet2_CVPR}$^{*}$ & $185$  & $73.2$ \\
				PWCNet \cite{Sun_2018_PWCNet_CVPR}$^{*}$ & $767$  & $72.1$ \\
				\hline
				Motion Vector & $>1000$  & $51.8$ \\
				Video Saliency \cite{Jiang_2018_Saliency_ECCV}$^{*}$ & $523$  & $56.9$ \\
				\hline
			\end{tabular}
		}
	}
    \end{minipage}%
    \begin{minipage}{0.5\textwidth}%
	\centering
	\subfigure{
	    \scalebox{0.85}{
	        \begin{tabular}{| l | c |}
				\hline
				Method	&  Accuracy ($\%$)		\\
				\hline	
				Motion-Aware  & $20.2$ \\
				\hline
				C3D \cite{Tran_C3D_2015_ICCV} & $23.0$ \\
                C3D + Motion-Aware & $\mathbf{26.1}$ \\
                \hline
                TCNN \cite{Hou_2017_TCNN_ICCV} & $28.4$ \\
                TCNN + Motion-Aware & $\mathbf{31.0}$ \\
                \hline
		    \end{tabular}
        }
    }
    \end{minipage}%
    \vspace{2ex}
	\caption{Left: Ablation study on motion representations, the input to our temporal augmented network. All speeds are evaluated on an image of resolution $112 \times 112$.  The speed only includes the time to compute the motion representation. $^{*}$ indicates the method uses a GPU for inference. Right: Anomalous activity recognition experiments. Our motion-aware feature can complement state-of-the-art video features and lead to large performance improvements. \label{tab:motion}}%
	\vspace{-4ex}
\end{table} 

\subsection{Anomalous Activity Recognition Experiments}

To further demonstrate the generalizability of our motion-aware feature, we use the same dataset to conduct anomalous action recognition experiments. Following the official setting, 
%we have 50 videos from each event and the training/testing ratio is 75/25. 
there are 4 splits and we report the average recognition accuracy. As can be seen in Table \ref{tab:motion} right, our motion-aware feature alone can achieve reasonable performance. When combined with other state-of-the-art video features, we can obtain large performance improvements, $3.1\%$ for C3D and $2.6\%$ for TCNN respectively. 

% \subsection{Per-class Breakdown}
% how to improve in the future 

\section{Conclusion}
\label{sec:conclusion}

In this work, we propose a temporal augmented network to learn a motion-aware feature. This feature alone can achieve competitive performance with previous state-of-the-art methods, and when combined with them, can achieve significant performance improvements. We also incorporate temporal context into the MIL ranking model by using an attention block. The learned attention weights can help to differentiate anomalous and normal video segments better. With the proposed motion-aware feature and temporal MIL ranking model, we achieve new state-of-the-art results for both anomaly detection and anomalous action recognition tasks in the UCF Crime dataset. 
Note that, our model still has difficulties in some known challenging scenarios, including fast motion, people grouping, low resolution, dark images, etc. In the future, we want to investigate other MIL formulations as in the recent zero-shot learning literature \cite{Zhu_2018_URL_CVPR}. We also want to make our pipeline have just one stage with end-to-end learning to obtain more robustness. 
% Video saliency is also an interesting direction to go. We want to explore other video saliency methods with consistent predictions. 

\paragraph{Acknowledegements}
We thank Amazon Web Service (AWS) for providing free EC2 access. We gratefully acknowledge the support of NVIDIA Corporation through the donation of the Titan Xp GPUs used in this work.

\bibliography{egbib}
\end{document}